\documentclass[letterpaper,10pt,conference]{ieeeconf}

\IEEEoverridecommandlockouts

\usepackage{graphicx}
\usepackage{tikz}
\usetikzlibrary{positioning,fit,calc,shapes.geometric}
\usepackage{adjustbox}
\usepackage[caption=false,font=footnotesize]{subfig}

\usepackage{booktabs}
\usepackage{array}
\usepackage{tabularx}
\usepackage{multirow}
\usepackage{diagbox}
\usepackage{adjustbox}
\usepackage{colortbl}

\usepackage{amsmath}
\usepackage{amssymb}
\usepackage{bbm}
\usepackage{dsfont}

\usepackage{algorithm}
\usepackage{algpseudocode}

\usepackage{xspace}
\usepackage{romannum}
\usepackage{textcomp}
\usepackage{siunitx}
\usepackage{soul}
\usepackage{xcolor}
\usepackage{xparse}
\usepackage{etoolbox}
\usepackage{lipsum}
\usepackage{balance}
\usepackage[font=small]{caption}

\usepackage{rpm_acronyms}
\usepackage{rpm_math}
\usepackage{rpm_misc}

\usepackage[hidelinks]{hyperref}
\usepackage{cleveref}

\crefname{figure}{Fig.}{Figs.}
\Crefname{figure}{Fig.}{Figs.}

\crefname{table}{Table}{Tables}
\Crefname{table}{Table}{Tables}

\crefname{section}{Section}{Sections}
\Crefname{section}{Section}{Sections}

\newcommand{\DATASETNAME}{Aria Office}
\newcommand{\METHODNAME}{FoundationVIO}
\newcommand{\best}[1]{\textbf{#1}}
\newcommand{\muted}[1]{\textcolor{gray}{#1}}

\begin{document}

\title{\LARGE \bf FoundationVIO: Enforcing Metric Scale\\ in Feed-forward Visual-Inertial Odometry}

\author{Jin Rhee, Yuhao Zhang, Frank Dellaert, and Maurice Fallon%
\thanks{
This project has been partly funded by the National Research Foundation of Korea (NRF) grant funded by the Korea government (MSIT)(No. RS-2024-00461409) and the UKRI Grant Mobile Robotic Inspector (EP/Z531212/1). For the purpose of open access, the authors have applied a Creative Commons Attribution (CC BY) license to any Accepted Manuscript version arising.
\textit{(Corresponding author: Jin Rhee.)}}%
\thanks{Jin Rhee, Yuhao Zhang, and Maurice Fallon are with Oxford Robotics Institute, University of Oxford, U.K. (e-mail: \texttt{jin.rhee@oxfordrobotics.institute}).}%
\thanks{Frank Dellaert is with School of Interactive Computing, Georgia Institute of Technology, USA (e-mail: \texttt{frank.dellaert@cc.gatech.edu}).}%
}

\maketitle

\bstctlcite{IEEEexample:BSTcontrol}

\begin{abstract}
Feed-forward geometric foundation models (GFMs) directly predict dense geometry and camera poses from batches of images. Emerging visual SLAM methods chain together consecutive GFM batches to form dense point cloud maps. A core challenge is how to address the ambiguous geometric scale of GFM predictions: independent predictions are locally consistent up-to-scale but not constrained by any direct physical measurements. Models trained on data with known scale improve the consistency of GFM reconstructions but again do not directly measure it. Consequently, trajectories and 3D reconstructions from GFM-enabled SLAM methods inherit this issue. Motivated by parallels with classical visual-inertial odometry, we introduce \METHODNAME{} which integrates inertial sensing with GFM predictions to recover metric scale reconstruction using a factor-graph based approach. We demonstrate odometry estimates competitive with classical state-of-the-art VIO methods and accurate dense reconstructions that outperform GFM-enabled SLAM methods. We also add a minimal GFM-based loop-closure mechanism to create a full Visual-Inertial SLAM system capable of creating globally accurate reconstructions and pose estimates. We evaluate our method on three datasets with well known sensor systems and demonstrate its compatibility on ubiquitous camera and IMU sensors. The method runs at approximately real-time on a consumer-grade laptop GPU. Code: \texttt{\textcolor{cyan}{\url{https://github.com/ori-drs/FoundationVIO}}}.
\end{abstract}

\section{Introduction}
\label{sec:introduction}

\begin{figure}[htb]
    \centering

    \begin{tikzpicture}
        \node[anchor=south west, inner sep=0] (img) at (0,0)
        {\includegraphics[width=\columnwidth]{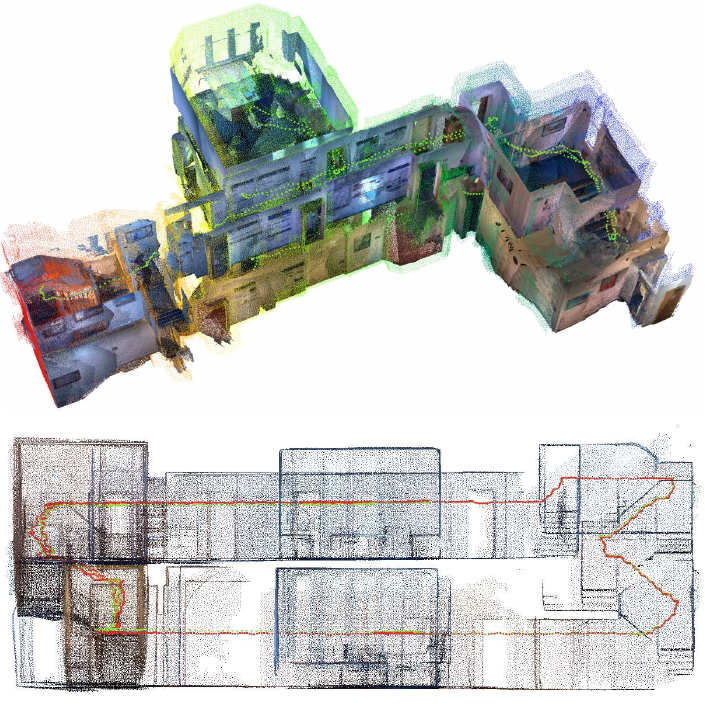}};
    \end{tikzpicture}

    \caption{Visualisation of the proposed system's trajectory and reconstruction using loop-closures on a self-collected sequence in a multi-floor office. \textit{Top:} Perspective view of the reconstruction with ground truth point cloud for reference. \textit{Bottom:} Side-view of the SLAM trajectory (green) against ground truth (red) and the downsampled reconstruction.}
    \label{fig:hero}
\end{figure}
Visual Simultaneous Localisation and Mapping (SLAM) is a fundamental and well-studied task in computer vision and robotics. Classical visual methods jointly estimate camera poses and a consistent sparse map using sparse feature extraction or photometric consistency~\cite{slam-handbook}. Recently developed feed-forward geometric foundation models (GFMs) can directly predict dense geometry and poses from a set of images. GFMs have spawned a new class of GFM-enabled visual SLAM methods~\cite{murai_mast3r-slam_2025, maggio_vggt-slam_2025} that incrementally chain independent feed-forward predictions to produce a consistent trajectory and dense reconstruction.

A key limitation when using GFMs has been the ambiguous geometric scale of their predictions. While a single feed-forward pass produces locally consistent geometry, its physical scale is unknown and varies between independent predictions. Single passes are limited by GPU memory to tens of images (for a mobile GPU). GFM-enabled visual SLAM methods~\cite{maggio_vggt-slam_2025, deng_vggt-long_2025} constrain drift by matching consecutive batch reconstructions and use optimisation to achieve self-consistent reconstructions in loop environments but cannot ground their output maps in physical scale. This presents a challenge for practical applications where consistent metric scale is critical, such as pose estimation, 3D reconstruction, or robot navigation.

The limitation is similar to the ambiguous scale problem in monocular visual SLAM. Some works have proposed to integrate IMU sensor measurements --- that are ubiquitous alongside cameras --- to help obtain metric-scale predictions~\cite{zhou_mast3r-fusion_2025, lozano_vi3_2026}. These methods demonstrated that GFM predictions can be refined with inertial measurements but these methods have been limited to early two-view GFMs or to grounding a single feed-forward prediction. Further research is needed to support multi-view GFMs and to more systematically combine inertial sensing and GFMs.

We propose \METHODNAME{} which integrates IMU measurements and multi-view GFM predictions into a factor-graph formulation to incrementally estimate poses and dense reconstruction. Our design is model-agnostic and we demonstrate that it works with several models - whose accuracy directly influences output reconstruction quality. Using the latest state-of-the-art model, \METHODNAME{} produces trajectory estimates that are competitive with well-established classical visual-inertial odometry methods on a range of self-collected and public datasets and outperforms existing inertial GFM approaches.
Our main contributions are as follows:
\begin{itemize}
    \item{We introduce a factor-graph based formulation to fuse IMU measurements with GFM predictions to produce a metric 3D reconstruction and an odometry pose estimate which relies on a custom relative scale factor between consecutive image batches.}
    \item{We minimally extend \METHODNAME{} into a SLAM system by adding a GFM-based loop-closure to produce globally accurate reconstructions and trajectories on three different datasets.}
    \item{We test our method with three different GFMs --- demonstrating that our system naturally profits from advances in GFM quality.}
\end{itemize}
We open-source our implementation of our method and self-collected dataset for future studies to evaluate against.
\section{Related Work}

\subsection{Sparse / Dense SLAM}
Classical visual SLAM systems rely on sparse geometric features and bundle adjustment to jointly estimate camera motion and map structure, achieving accurate and robust localisation~\cite{campos_orb-slam3_2021}. Direct methods instead exploit photometric consistency between images, reducing reliance on explicit feature extraction~\cite{engel_dso_2017}. Dense SLAM extends beyond sparse landmarks by reconstructing detailed scene geometry often by relying on active depth sensors~\cite{newcombe_kinectfusion_2011, whelan_elasticfusion_2015}. More recent learning-based approaches instead use learned dense correspondence estimation or learned scene representations to support monocular dense SLAM without requiring these specialised depth sensors. DROID-SLAM~\cite{teed_droid-slam_2021} estimates dense inter-frame correspondences and iteratively refines camera poses through recurrent updates and differentiable dense bundle adjustment, while neural implicit methods such as NICE-SLAM~\cite{zhu_nice-slam_2022} jointly optimise camera poses and implicit scene representations to create dense reconstructions.

\subsection{Feed-Forward Reconstruction Models (Geometric Foundation Models)}
Recent advances in geometric foundation models (GFMs) enable feed-forward dense point cloud reconstruction and camera pose estimation directly from multi-view monocular images. The first two-view feed-forward reconstruction models, DUSt3R~\cite{wang_dust3r_2024} and MASt3R~\cite{leroy_mast3r_2024}, predict dense point cloud reconstructions from two images. This approach has evolved to directly reconstruct using a batch of images~\cite{wang_spann3r_2025}, to predict camera poses using a prediction head~\cite{yang_fast3r_2025, wang_vggt_2025}, and to accept camera intrinsic parameters as an optional input~\cite{keetha_mapanything_2025, lin_depth_2025}.
While individual GFM predictions are reasonably consistent for a small batch of overlapping images, consistency between independent feed-forward passes is unconstrained.
Most notably, the scale of these predictions can drift, similar to the problem of unobservable scale in classical monocular visual SLAM.
A parallel line of work adapts GFMs to streaming input by maintaining a latent memory as a recurrent state~\cite{wang_cut3r_2025}, causal KV cache~\cite{zhuo_streamingvggt_2026}, or learned attention contexts~\cite{chen_lingbotmap_2026}. However, as memory is a learned latent rather than a geometric state, there is no straightforward way to use an optimisation based method to propagate later evidence such as loop-closures into earlier predictions; drift therefore accumulates and is never explicitly corrected~\cite{cheng_longstream_2026, chen_lingbotmap_2026}.

\subsection{SLAM with Geometric Foundation Models}
To extend multi-view GFMs to handle streaming image sequences and enable explicit correction of drift, recent approaches incorporate SLAM techniques into GFMs to carry out incremental reconstruction, map optimisation, and loop closure.
MASt3R-SLAM~\cite{murai_mast3r-slam_2025} extends the two-view MASt3R~\cite{leroy_mast3r_2024} model into a real-time dense SLAM system by incrementally aligning pointmaps and optimising a global $\mathrm{Sim}(3)$ graph. ViSTA-SLAM~\cite{zhang_vista-slam_2026} also employs a learned two-view reconstruction frontend with $\mathrm{Sim}(3)$ pose-graph optimisation. Using the multi-view VGGT model, VGGT-SLAM~\cite{maggio_vggt-slam_2025} incrementally constructs local submaps and successively registers them through optimisation on the $\mathrm{SL}(4)$ manifold. AIM-SLAM~\cite{jeon_aim-slam_2026} adaptively selects informative multi-view keyframes and jointly optimises their alignment.

These systems primarily target visual consistency and remain subject to the scale ambiguity of monocular reconstruction. To achieve real-world scale consistency, recent works study how to use inertial sensing to metrically anchor GFM predictions. MASt3R-Fusion~\cite{zhou_mast3r-fusion_2025} tightly fuses IMU measurements with pairwise MASt3R predictions using a factor graph to recover metric scale. It further utilises image feature tokens from MASt3R's encoder to obtain loop-closure candidates. One approach to obtaining better depth prediction is to use more recent multi-view models. In recent work, VI3~\cite{lozano_vi3_2026} produces metric-scale GFM predictions by deriving a metric motion reference from IMU preintegration to condition compatible GFMs or to refine outputs after main inference. VI3 limits its method to grounding single-batch predictions on short image sequences, a different class of problem to the incremental nature of SLAM. Our method generalises visual-inertial GFM mapping to multi-view models while incrementally fusing successive batches over long sequences, enabling metric-scale reconstruction beyond the pairwise and short-window settings considered in prior work.
\section{Method}
\begin{figure}[htb]
    \centering

    \begin{tikzpicture}
        \node[anchor=south west, inner sep=0] (img) at (0,0)
        {\includegraphics[width=\columnwidth]{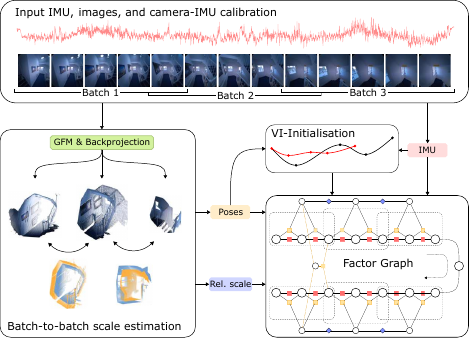}};
    \end{tikzpicture}

    \caption{Component diagram of the proposed system. The factor graph contains per-batch scale $l_k=\log{s_k}$ and per-frame pose $\mathbf{T}_i$, velocity $\mathbf{v}_i$, and biases $\mathbf{b}_i$ as variables. IMU data, up-to-scale GFM pose predictions, and the batch-to-batch relative scale estimates are used as measurements for the IMU preintegration factor (red), GFM pose factor (yellow), and scale consistency factor (blue).
    }
    \label{fig:method}
\end{figure}
\begin{figure}[htb]
    \centering

    \begin{tikzpicture}
        \node[anchor=south west, inner sep=0] (img) at (0,0)
        {\includegraphics[width=\columnwidth]{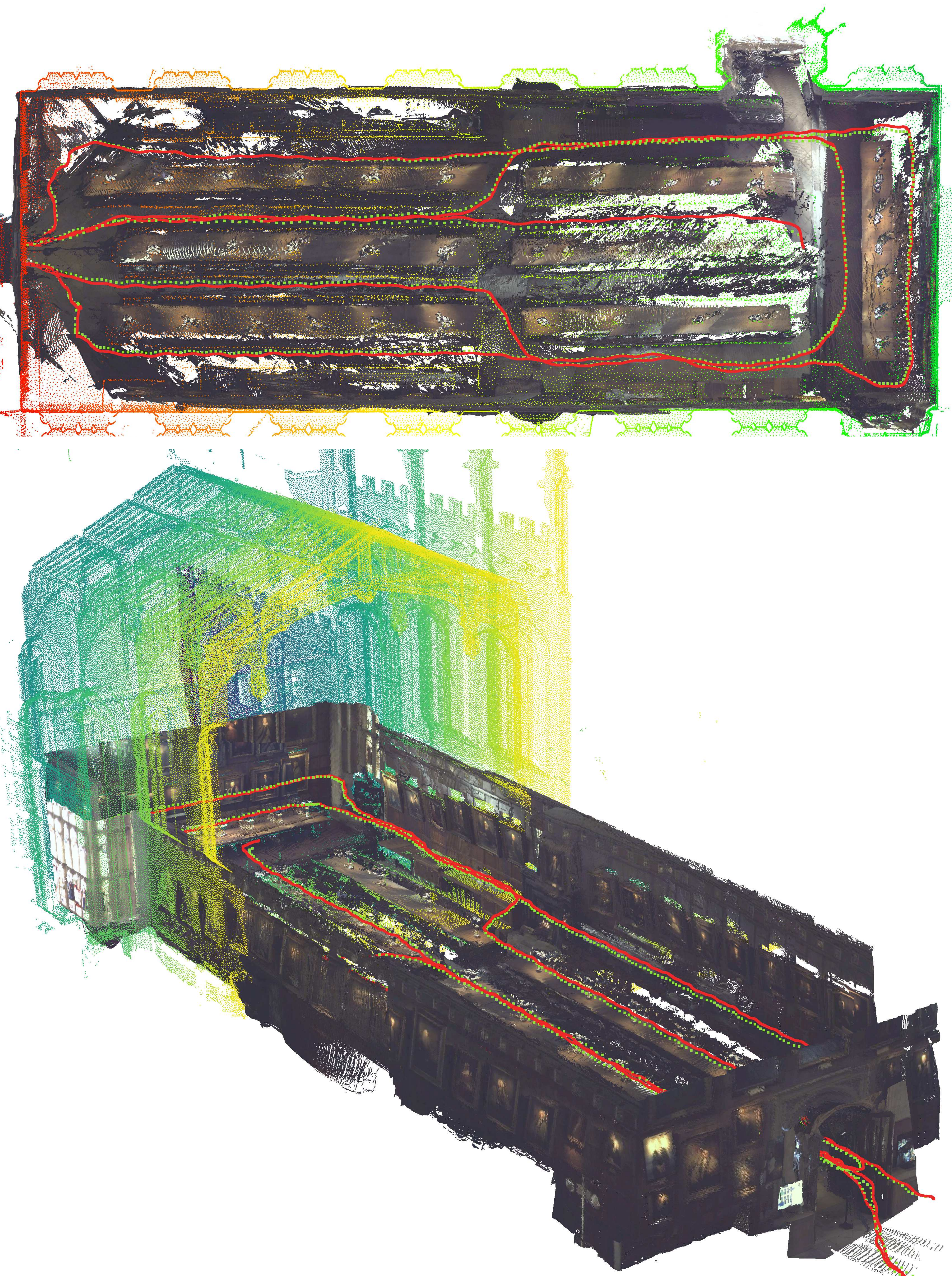}};
    \end{tikzpicture}

    \caption{Top-down and perspective views of estimated SLAM trajectory and reconstruction on Christ-Church-03 sequence from Oxford Spires. Ground truth trajectory in red and TLS scan is visualised for reference.}
    \label{fig:ch-ch-03-pretty}
\end{figure}
An overview of the proposed method is shown in Fig.~\ref{fig:method}. We propose a factor-graph based design that takes IMU measurements, up-to-scale poses from GFM predictions, and relative scale measurements derived between consecutive predictions in order to create a natively metric scale trajectory and 3D reconstruction. GFMs such as MapAnything~\cite{keetha_mapanything_2025} or DA3~\cite{lin_depth_2025} have a degree of real-world scale by training on datasets with known scale --- however this merely improves their predictive capability and predictions are not constrained by physical measurements. Our method recovers real-world scale --- as demonstrated in Fig. \ref{fig:ch-ch-03-pretty} --- using IMU sensor measurements and is compatible with sensors typically used on drones, phones or AR glasses.

\subsection{Notation and preliminaries}
We denote single object vectors or matrices in bold (e.g. image \textbf{I} as an array of RGB values) and sets of objects in calligraphic font (e.g., set of images $\mathcal{I}$). A batch consists of a set of frames, where each frame corresponds to an input image \textbf{I}, pose \textbf{T}, depth map \textbf{D}, and confidence map \textbf{C}.

In the results section we use models from Depth Anything 3 (DA3)~\cite{lin_depth_2025} and VGGT-$\Omega$~\cite{wang_vggt-omega_2026} though any model that jointly infers predictions from a set of input images can be used~\cite{wang_vggt_2025, keetha_mapanything_2025}. From model predictions, per-frame geometry can be obtained by backprojecting the depth map \textbf{D} with the camera intrinsics obtained either from a prior calibration or directly from model predictions.

\subsection{Visual-inertial initialisation}
\label{subsec:vi_initialisation}
We begin each sequence with a visual-inertial initialisation phase aided by GFM predictions, similar to MASt3R-Fusion~\cite{zhou_mast3r-fusion_2025}. Specifically, we adapt the visual-inertial initialisation method from VINS-Mono~\cite{qin_vins-mono_2018}. This module uses Structure-from-Motion (SfM) to estimate the gyroscope biases and the gravity vector by aligning IMU preintegration to the SfM poses.
We swap that SfM step with camera pose predictions from a GFM. This allows robust initialisation under conventionally degenerate conditions such as low texture or small parallax and negates the need for heuristic checks of non-zero parallax or verification of SfM pose quality with feature correspondences. Instead, we maintain a simpler fixed sliding window of frames until IMU excitation, computed as the standard deviation of the accelerometer norm, exceeds a threshold which indicates motion. The accepted window of frames is then fed to the GFM to produce an up-to-scale trajectory, onto which we then fit IMU preintegration to estimate initial conditions (such as the gravity vector and gyroscope bias). Initial estimates from the visual-inertial initialisation are used to bootstrap the factor graph formulation and to avoid IMU divergence.

\subsection{Model inference}
\label{subsec:model_inference}
As images are received by the algorithm, we group them into overlapping batches $\mathcal{I}_{k}=\{\mathbf{I}^{k}_{1}, \dots,\mathbf{I}^{k}_{B}\}$ as seen at the top of Fig. \ref{fig:method}. For each new batch, the first image is chosen to overlap with the last image of the previous batch. We feed the image batch into the GFM, obtaining a set of initial camera pose estimates $\mathcal{T}_{k}=\{\mathbf{T}^{k}_{1},\dots,\mathbf{T}^{k}_{B}\}$ and pointmaps $\mathcal{P}_{k}=\{\mathbf{P}^{k}_{1},\dots,\mathbf{P}^{k}_{B}\}$ that are self-consistent up-to-scale, as well as a set of confidence maps $\mathcal{C}_{k}=\{\mathbf{C}^{k}_{1},\dots,\mathbf{C}^{k}_{B}\}$. As the predictions use the first camera pose as the reference frame, we transform the poses and pointmaps into each pose's IMU body frame using the known calibrated extrinsics $\mathbf{X}_{body}^{imu}$.
In the next step we mask out points from the model predictions that are likely to contribute to inaccurate geometry. First, we filter out points whose depth exceeds a threshold multiple of the trajectory baseline. We then generate a depth-based edge mask to filter out points that represent spurious noise present on an edge between a foreground object and the background as inspired by~\cite{keetha_mapanything_2025}.

\subsection{Batch-to-batch relative scale estimation}
\label{subsec:irls}
Given independent pointmap predictions from batches, we refine the consistency of consecutive batches by robustly estimating their relative geometric scale as seen in the lower left section of Fig. \ref{fig:method}. Specifically, for the image $\mathbf{I}_{overlap}$ shared by consecutive batches $k-1$ and $k$ we have access to two separate pointmap and confidence-map predictions, $\{\mathbf{P}^{k-1}_{B}, \mathbf{C}^{k-1}_{B}\}$ and $\{\mathbf{P}^{k}_{0}, \mathbf{C}^{k}_{0}\}$. These pointmaps are aligned by estimating a $\mathrm{Sim}(3)$ transformation using an IRLS optimisation scheme~\cite{deng_vggt-long_2025}. At each iteration, the optimisation solves a confidence-weighted Umeyama alignment between the two pointmaps. The estimated scale component, $s_k^{\mathrm{rel}}$, defines the relative scale of batch $k$ with respect to batch $k-1$.

\subsection{Factor graph construction}
\label{subsec:factor_graph}

Given the up-to-scale poses $\mathcal{T}_{k}$, relative batch scale $s_{k}^{\mathrm{rel}}$, and IMU measurements $\mathcal{U}_{ij} = \{\tilde{\mathbf{a}}_t, \tilde{\boldsymbol{\omega}}_t\}_{t=i}^{j}$, we construct a factor graph consisting of per-image variables $\mathcal{X}_{i}=\{\mathbf{T}_{i}, \mathbf{v}_{i},\mathbf{b}_{i}\}$ and per-batch variables $l_{k}$ where $\mathbf{T}_{i} = \{\mathbf{t}_{i},\mathbf{R}_{i}\}\in\mathrm{SE}(3)$ is the $i$-th body pose, $\mathbf{v}_{i}$ is the $i$-th body velocity, $\mathbf{b}_{i}=\{\mathbf{b}_{i}^{a}\in\mathbb{R}^{3},\mathbf{b}_{i}^{g}\in\mathbb{R}^{3}\}$ is the $i$-th IMU bias variable, and $l_{k} = \log{s_k}$ is the $k$-th batch scale variable in log space. The batch scale variable represents the scale of the pointmaps and poses relative to metric scale and is stored in log space to simplify the multiplication of relative batch scales as additions.

We use the IMU preintegration factor~\cite{forster_imu_2015} whose residual is defined as
\begin{align*}
\mathbf{r}_{ij}^{IMU}(\mathcal{X}_{i},\mathcal{X}_{j}) = \begin{bmatrix}
\mathbf{R}_i^T(\mathbf{t}_j - \mathbf{t}_i - \mathbf{v}_i\Delta t - \frac{1}{2}\mathbf{g}\Delta t^2) - \hat{\mathbf{t}}_{ij}\\
\mathbf{R}_i^T(\mathbf{v}_j - \mathbf{v}_i - \mathbf{g}\Delta t) - \hat{\mathbf{v}}_{ij}\\
\text{Log}(\hat{\mathbf{R}}_{ij}^T\mathbf{R}_i^T\mathbf{R}_j)\\
\mathbf{b}_j^a - \mathbf{b}_i^a\\
\mathbf{b}_j^g - \mathbf{b}_i^g
\end{bmatrix}
\end{align*}
where $\Delta t$ is the preintegration time, $\mathbf{g}$ is the gravity vector, and $\hat{\mathbf{t}}_{ij}, \hat{\mathbf{v}}_{ij}, \hat{\mathbf{R}}_{ij}$ are the preintegration terms.

To utilise the up-to-scale pose predictions in a $\mathrm{SE}(3)$ factor graph, we implement a custom-defined factor that connects two consecutive poses $\mathbf{T}_{i}$ and $\mathbf{T}_{j}$ with the corresponding batch scale variable $l_k = \log{s_{k}}$.
It takes as measurement the up-to-scale relative pose from the feed-forward model, $\mathbf{Z}_{i,j} = \mathbf{T}_j^{-1}\mathbf{T}_{i} = \{\mathbf{Z}_t, \mathbf{Z}_R\}$.

The residual is defined as
\begin{align}
    \mathbf{r}_{i,j}^{GFM}(\mathbf{T}_{i}, \mathbf{T}_{j}, s_{k})&=
    \begin{bmatrix}
        \mathbf{r}_R\\ \mathbf{r}_t
    \end{bmatrix} =
    \begin{bmatrix}
        \text{Log}(\mathbf{Z}_R^T\mathbf{R}_i^T\mathbf{R}_j) \\ \mathbf{Z}_t - s_k \mathbf{R}_i^T(\mathbf{t}_j-\mathbf{t}_i)
    \end{bmatrix}
\end{align}
where $\text{Log}(\cdot)$ denotes the $\mathrm{SO}(3)$ logarithmic map. Note that this is identical to a relative pose factor, with the exception of the batch scale $s_k$ being applied to bring the translational residual $\mathbf{r}_{t}$ into batch scale.

To ensure consistency of scale across batches, we implement a simple binary factor whose edges connect to consecutive batch scale variables. It takes in the relative scale ratio measurement $s_{k+1}^{\mathrm{rel}}$. Its residual is simply defined as
\begin{equation}
    \mathbf{r}_k^{scale}(l_{k}, l_{k+1}) = (l_{k+1} - l_k) - \log{s_{k+1}^{\mathrm{rel}}}
\end{equation}
where we use subtraction in log-scale to simplify multiplication.

\subsection{Factor graph optimisation}
\label{subsec:opt_factor_graph}
We optimise the constructed factor graph using the iSAM2 algorithm~\cite{kaess_isam2_2012} implemented in GTSAM~\cite{gtsam}. Unlike a fixed sliding window, iSAM2 relinearises variables whose linear approximation error grows beyond a threshold. We call new Gauss-Newton update steps whenever a new set of factors are added from a new batch.
Once a sequence ends and the factor graph has been fully optimised, each batch's scale variable is used to rescale its point cloud by a simple multiplication and is pinned on the batch's first pose to create a consistent metric reconstruction.

\subsection{Loop-closures}
\label{subsec:loop_closures}
To extend the system from a purely odometry system into a SLAM system capable of detecting loop-closures and correcting for drift, we minimally extend our factor graph by following an approach similar to VGGT-SLAM~\cite{maggio_vggt-slam_2025}. \textit{This SLAM extension is not the main focus of our work}. At each frame, we use the SALAD~\cite{izquierdo_salad_2024} global descriptor to generate an embedding matrix. We ignore recent batches to prioritise temporally distant loop-closure candidates and select the candidate frame with the lowest L2-norm.
We create a loop-closure batch consisting of the query and target frame to obtain GFM predictions in the same manner as with normal batches. From the loop-closure batch we perform the same batch-to-batch scale estimation in Section~\ref{subsec:irls} against both the target and query batch. Finally, we add the batch into the factor graph, connecting the loop-closure batch's scale node to others with a robust Huber loss kernel to dampen the effect of incorrect loop-closure detections.
\section{Evaluation}
\label{sec:evaluation}
We evaluate both the trajectory and reconstruction quality of \METHODNAME{} across three datasets.
We first compare \METHODNAME{}'s pose estimation accuracy with respect to both classical and GFM-enabled methods. Results are presented both with the system configured in odometry mode and also in SLAM mode with loop-closures enabled.
We then evaluate \METHODNAME{}'s dense 3D reconstructions against other state-of-the-art GFM-enabled SLAM methods. Finally, we evaluate the real-time performance of our method. All experiments are carried out on a laptop equipped with an Intel Core(TM) i7-13850HX and an RTX 3500 Ada Generation Laptop GPU with 12\,GB VRAM for equal comparison.

\subsection{Datasets}
The three datasets consist of 1) a self-collected indoor dataset, 2) the Oxford Spires dataset~\cite{tao_spires_2025}, and 3) the ScaRF-SLAM dataset~\cite{zhang_scarf_2026}. Together, these datasets span different indoor and outdoor scenes and cover a wide range of sensor rigs and environments.

\textbf{The \DATASETNAME{} dataset} was self-collected using the Meta Aria glasses (Gen 1) which is equipped with two large-FoV greyscale cameras (HFoV$\,=150^{\circ}$), a front-facing RGB camera (FoV$\,=110^{\circ}$, 15\,Hz at 1408\,x\,1408 resolution), and two IMUs (1\,kHz, 800\,Hz). The sequences consist of head-worn trajectories touring several floors of an office environment in multiple loops and pass through narrow corridors, low-texture staircases, and office rooms.
In our experiments, we use only pinhole rectified images (98.2$^{\circ}$$\times$98.2$^{\circ}$) from the forward-facing camera and the 1\,kHz IMU located near the right ear. The device's camera intrinsics and camera-to-IMU extrinsics are independently calibrated using Kalibr~\cite{rehder_extending_2016, el-sheimy_analysis_2008}. We compare performance to reference trajectories computed using a post-processing SLAM called MPS provided by the manufacturers which has been demonstrated to achieve extremely low drift~\cite{lamaria}.
So as to be able to quantitatively analyse reconstruction accuracy, we created a high-precision terrestrial laser scan (TLS) of the building using a Leica BLK360 scanner. Reference trajectories are aligned to the accurate TLS scan via a rigid transformation.

\textbf{The Oxford Spires dataset}~\cite{tao_spires_2025} contains long sequences captured with a custom perception unit, using Alphasense cameras (20\,Hz at 1440\,x\,1080 resolution) and IMUs at 400\,Hz. It features indoor-outdoor transitions around landmarks. We choose test sequences that contain both indoor and outdoor environments and ground truth trajectories.

\textbf{The ScaRF-SLAM dataset}~\cite{zhang_scarf_2026} consists of sequences in a multi-floor office environment captured with an Insta360 ONE RS 1-inch, containing dual fisheye cameras (30\,Hz at 1472\,x\,1440 resolution) and IMU at 1\,kHz. The sequences also include an initialisation period where the device is lifted gently to aid visual-inertial initialisation.
Similarly to \DATASETNAME{}, we use pinhole rectified images only of the forward facing camera (Oxford Spires at 91.6$^\circ$$\times$75.4$^\circ$ and ScaRF-SLAM at 81.5$^\circ$$\times$ 65.9$^\circ$) and the IMU sensor. Both datasets contain a millimetre-accurate map from a terrestrial laser scanner and ground truth trajectories obtained by aligning mobile LiDAR scans (fixed to the camera) to the laser map.

\newsavebox{\trajBox}
\newlength{\trajW}
\newlength{\trajH}

\begin{figure}[htb]
    \centering
    \setlength{\trajW}{\columnwidth}
    \sbox{\trajBox}{\includegraphics[width=\trajW]{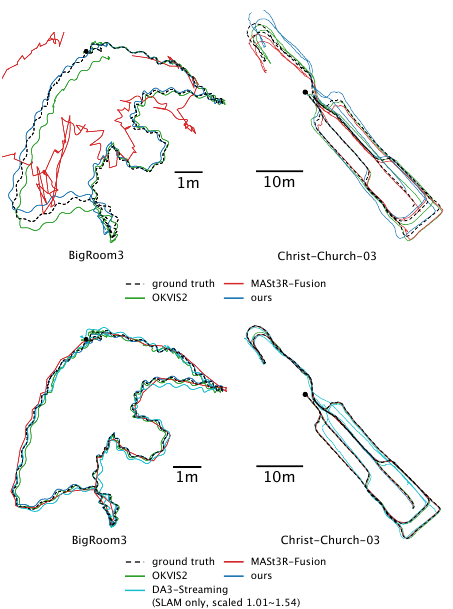}}%
    \setlength{\trajH}{\ht\trajBox}
    \newcommand{\trajB}[2]{%
        \begin{tikzpicture}
            \clip (0, #1\trajH) rectangle (\trajW, #2\trajH);
            \node[anchor=south west, inner sep=0] at (0,0) {\usebox{\trajBox}};
        \end{tikzpicture}}

    \subfloat[Odometry Poses (no loop-closure)]{\trajB{0.5}{1}\label{fig:traj:odom}}

    \vspace{2pt}

    \subfloat[SLAM Poses (with loop-closure)]{\trajB{0}{0.5}\label{fig:traj:slam}}

    \caption{Odometry and SLAM trajectories of OKVIS2, MASt3R-Fusion, \METHODNAME{}, and DA3-Streaming on a variety of sequences. Each sequence's start is denoted by a black dot. The outlier poses used for visual-inertial initialisation in MASt3R-Fusion are removed for fair comparison.}
    \label{fig:traj_comparison}
\end{figure}
\subsection{Pose estimate accuracy as an odometry system}

To evaluate our method's odometry pose estimate, we compute the root mean squared error (RMSE) of its relative pose error (RPE) against the reference trajectory.
To demonstrate and evaluate \METHODNAME{}'s general compatibility with other GFMs, we include results using VGGT-$\Omega$~\cite{wang_vggt-omega_2026} (1.0B parameters) and two sizes of DA3~\cite{lin_depth_2025} (Large 0.3B parameters, Nested 1.4B parameters).
We compare our method against classical visual-inertial methods including OKVIS2~\cite{leutnegger_okvis2_2022}, ORB-SLAM3~\cite{campos_orb-slam3_2021}, and VINS-Fusion~\cite{qin_vins-mono_2018}. We also compare against MASt3R-Fusion~\cite{zhou_mast3r-fusion_2025} which fuses IMU measurements with MASt3R~\cite{leroy_mast3r_2024}, an early two-view GFM model. MASt3R-Fusion is tested with its default input rate configuration, while classical methods are tested using the full image sequences. In order to attain near real-time performance and fit laptop GPU memory constraints, we test \METHODNAME{} with image sequences downsampled to 5\,Hz and use a batch size of 10 images. The best performance in the classical and GFM-enabled methods are highlighted independently to compare the class of GFM-enabled methods against classical methods. 

\begin{table*}[t!]
\centering
\scriptsize
\caption{RMSE of RPE on all datasets with per-dataset means. Best and second-best are marked within the classical and GFM groups separately. All methods use rectified pinhole images with the same FoV.}
\setlength{\tabcolsep}{2.5pt}
\begin{adjustbox}{max width=\textwidth}
\begin{tabular}{l ccccccccc c ccccc c ccc c}
\toprule
 & \multicolumn{10}{c}{\DATASETNAME{}} & \multicolumn{6}{c}{ScaRF-SLAM} & \multicolumn{4}{c}{Oxford Spires} \\
\cmidrule(lr){2-11} \cmidrule(lr){12-17} \cmidrule(lr){18-21}
 & AT1 & AT2 & BR2 & BR3 & BR4 & NT1 & NT2 & NT3 & NT4 & Mean
 & r01 & r02 & r03 & r04 & r05 & Mean
 & CC02 & CC03 & CC05 & Mean \\
\midrule
\multicolumn{21}{c}{\textbf{Trans.\ RPE} (m/m) $\downarrow$} \\[1pt]
OKVIS2$^\dagger$
& \muted{\best{0.046}} & \muted{\best{0.063}} & \muted{\best{0.057}} & \muted{0.051} & \muted{\underline{0.053}} & \muted{\underline{0.060}} & \muted{\best{0.046}} & \muted{\underline{0.065}} & \muted{\best{0.181}} & \muted{\underline{0.069}}
& \muted{\best{0.030}} & \muted{\best{0.027}} & \muted{\best{0.026}} & \muted{\best{0.036}} & \muted{\best{0.035}} & \muted{\best{0.031}}
& \muted{0.095} & \muted{0.100} & \muted{0.083} & \muted{0.093} \\
ORB-SLAM3$^\dagger$
& \muted{$\times$} & \muted{\underline{0.064}} & \muted{$\times$} & \muted{\best{0.011}} & \muted{\best{0.009}} & \muted{\best{0.021}} & \muted{$\times$} & \muted{\best{0.025}} & \muted{$\times$} & \muted{\best{0.026$^{*}$}}
& \muted{\underline{0.044}} & \muted{\underline{0.033}} & \muted{\underline{0.032}} & \muted{\underline{0.058}} & \muted{0.076} & \muted{\underline{0.049}}
& \muted{\underline{0.031}} & \muted{\best{0.018}} & \muted{\best{0.039}} & \muted{\underline{0.029}} \\
VINS-Fusion$^\dagger$
& \muted{$\times$} & \muted{$\times$} & \muted{\underline{1.871}} & \muted{\underline{0.035}} & \muted{0.053} & \muted{1.284} & \muted{\underline{1.214}} & \muted{4.901} & \muted{$\times$} & \muted{1.560$^{*}$}
& \muted{0.127} & \muted{0.058} & \muted{0.066} & \muted{0.389} & \muted{\underline{0.044}} & \muted{0.137}
& \muted{\best{0.020}} & \muted{\underline{0.019}} & \muted{\underline{0.041}} & \muted{\best{0.026}} \\
\midrule
MASt3R-Fusion
& 0.142 & 0.609 & 0.480 & $\times$ & 0.694 & 0.152 & 0.147 & 0.107 & 0.116 & 0.306$^{*}$
& 0.098 & 0.097 & 0.119 & 0.089 & 0.151 & 0.111
& \underline{0.236} & {0.165} & \underline{0.476} & \underline{0.292} \\
\METHODNAME{} (DA3-L)
& 0.134 & 0.156 & 0.181 & 0.179 & 0.176 & 0.119 & 0.131 & \underline{0.088} & \underline{0.088} & 0.139
& \underline{0.059} & 0.131 & \underline{0.079} & 0.095 & 0.103 & 0.093
& $\times$ & \underline{0.153} & 0.160 & 0.157$^{*}$ \\
\METHODNAME{} (VGGT-$\Omega$)
& \underline{0.107} & \underline{0.103} & \underline{0.132} & \underline{0.136} & \underline{0.123} & \underline{0.103} & \underline{0.089} & 0.088 & 0.089 & \underline{0.108}
& 0.102 & \underline{0.094} & 0.091 & \underline{0.061} & \underline{0.060} & \underline{0.082}
& 0.301 & 0.154 & 0.156 & 0.204 \\
\METHODNAME{} (DA3-N)
& \best{0.037} & \best{0.033} & \best{0.028} & \best{0.027} & \best{0.028} & \best{0.041} & \best{0.039} & \best{0.036} & \best{0.036} & \best{0.034}
& \best{0.034} & \best{0.031} & \best{0.034} & \best{0.043} & \best{0.045} & \best{0.037}
& \best{0.035} & \best{0.044} & \best{0.045} & \best{0.041} \\
\midrule
\multicolumn{21}{c}{\textbf{Rot.\ RPE} ($^{\circ}$/m) $\downarrow$} \\[1pt]
OKVIS2$^\dagger$
& \muted{\best{0.67}} & \muted{\underline{0.63}} & \muted{\underline{0.79}} & \muted{0.36} & \muted{0.38} & \muted{\underline{0.45}} & \muted{\best{0.74}} & \muted{\underline{0.63}} & \muted{\best{4.51}} & \muted{\underline{1.02}}
& \muted{\best{0.29}} & \muted{\underline{0.31}} & \muted{\underline{0.34}} & \muted{\best{0.33}} & \muted{\underline{0.24}} & \muted{\underline{0.30}}
& \muted{0.20} & \muted{0.54} & \muted{\underline{0.21}} & \muted{0.31} \\
ORB-SLAM3$^\dagger$
& \muted{$\times$} & \muted{\best{0.43}} & \muted{$\times$} & \muted{\underline{0.30}} & \muted{\underline{0.35}} & \muted{\best{0.34}} & \muted{$\times$} & \muted{\best{0.39}} & \muted{$\times$} & \muted{\best{0.36$^{*}$}}
& \muted{\underline{0.31}} & \muted{\best{0.27}} & \muted{\best{0.29}} & \muted{\underline{0.33}} & \muted{\best{0.20}} & \muted{\best{0.28}}
& \muted{\underline{0.12}} & \muted{\best{0.14}} & \muted{0.22} & \muted{\underline{0.16}} \\
VINS-Fusion$^\dagger$
& \muted{$\times$} & \muted{$\times$} & \muted{\best{0.61}} & \muted{\best{0.25}} & \muted{\best{0.29}} & \muted{3.17} & \muted{\underline{3.91}} & \muted{3.78} & \muted{$\times$} & \muted{2.00$^{*}$}
& \muted{2.33} & \muted{1.19} & \muted{1.09} & \muted{3.24} & \muted{0.75} & \muted{1.72}
& \muted{\best{0.11}} & \muted{\underline{0.15}} & \muted{\best{0.11}} & \muted{\best{0.12}} \\
\midrule
MASt3R-Fusion
& 0.43 & 0.71 & 0.58 & $\times$ & \best{0.55} & \best{0.50} & \best{0.50} & 0.49 & \best{0.36} & 0.51$^{*}$
& 0.56 & 0.91 & 0.60 & 0.62 & 0.55 & 0.65
& \best{0.17} & \best{0.22} & \best{0.14} & \best{0.18} \\
\METHODNAME{} (DA3-L)
& 0.36 & 0.44 & 0.55 & 0.54 & 0.62 & 0.57 & 0.58 & 0.45 & 0.44 & 0.51
& 0.38 & 0.43 & \best{0.40} & \underline{0.50} & \best{0.30} & \best{0.40}
& $\times$ & 0.28 & 0.26 & 0.27$^{*}$ \\
\METHODNAME{} (VGGT-$\Omega$)
& \underline{0.36} & \underline{0.44} & \underline{0.55} & \underline{0.53} & 0.62 & \underline{0.56} & 0.57 & \underline{0.45} & 0.44 & \underline{0.50}
& \best{0.37} & \underline{0.43} & 0.45 & \best{0.48} & 0.32 & 0.41
& 0.25 & 0.27 & 0.26 & 0.26 \\
\METHODNAME{} (DA3-N)
& \best{0.34} & \best{0.43} & \best{0.52} & \best{0.51} & \underline{0.59} & 0.56 & \underline{0.56} & \best{0.44} & \underline{0.43} & \best{0.49}
& \underline{0.37} & \best{0.42} & \underline{0.41} & 0.50 & \underline{0.31} & \underline{0.40}
& \underline{0.22} & \underline{0.26} & \underline{0.25} & \underline{0.24} \\
\bottomrule
\end{tabular}
\end{adjustbox}
\par\vspace{2pt}
{\raggedright
\footnotesize
$\times$: diverged run ($>5$\,m/m or $>5^{\circ}$/m). $^\dagger$Loop closure disabled. $^{*}$Mean over non-diverged sequences only.\par}
\label{tab:odom_eval_all}
\end{table*}
\begin{table*}[t!]
\centering
\scriptsize
\caption{RMSE of APE (m) $\downarrow$ after full SLAM on all datasets, with per-dataset means. Best and second-best are marked within the classical and GFM groups separately. All methods use rectified pinhole images with the same FoV.}
\setlength{\tabcolsep}{2.5pt}
\begin{adjustbox}{max width=\textwidth}
\begin{tabular}{l ccccccccc c ccccc c ccc c}
\toprule
 & \multicolumn{10}{c}{\DATASETNAME{}} & \multicolumn{6}{c}{ScaRF-SLAM} & \multicolumn{4}{c}{Oxford Spires} \\
\cmidrule(lr){2-11} \cmidrule(lr){12-17} \cmidrule(lr){18-21}
 & AT1 & AT2 & BR2 & BR3 & BR4 & NT1 & NT2 & NT3 & NT4 & Mean
 & r01 & r02 & r03 & r04 & r05 & Mean
 & CC02 & CC03 & CC05 & Mean\\
\midrule
OKVIS2
& \muted{\best{0.165}} & \muted{\best{0.204}} & \muted{\best{0.087}} & \muted{0.133} & \muted{\underline{0.085}} & \muted{\best{0.260}} & \muted{\best{0.151}} & \muted{\underline{0.453}} & \muted{\best{1.442}} & \muted{\best{0.331}}
& \muted{\best{0.117}} & \muted{\best{0.096}} & \muted{\best{0.077}} & \muted{\best{0.182}} & \muted{\underline{0.107}} & \muted{\best{0.116}}
& \muted{\underline{1.597}} & \muted{0.299} & \muted{\best{0.630}} & \muted{\underline{0.842}}\\
ORB-SLAM3
& \muted{$\times$} & \muted{\underline{0.357}} & \muted{$\times$} & \muted{\best{0.026}} & \muted{\best{0.041}} & \muted{\underline{4.363}} & \muted{\underline{1.944}} & \muted{\best{0.202}} & \muted{\underline{3.974}} & \muted{\underline{1.558$^{*}$}}
& \muted{\underline{0.322}} & \muted{\underline{0.127}} & \muted{\underline{0.116}} & \muted{\underline{0.294}} & \muted{0.301} & \muted{\underline{0.232}}
& \muted{1.931} & \muted{\best{0.192}} & \muted{$\times$} & \muted{1.061$^{*}$}\\
VINS-Fusion
& \muted{\underline{4.863}} & \muted{$\times$} & \muted{$\times$} & \muted{\underline{0.083}} & \muted{0.169} & \muted{$\times$} & \muted{$\times$} & \muted{$\times$} & \muted{$\times$} & \muted{1.705$^{*}$}
& \muted{0.614} & \muted{0.127} & \muted{0.150} & \muted{1.129} & \muted{\best{0.101}} & \muted{0.424}
& \muted{\best{0.708}} & \muted{\underline{0.253}} & \muted{\underline{0.858}} & \muted{\best{0.607}}\\
\midrule
\rowcolor{gray!15}
MASt3R-Fusion (BA + LC)
& \underline{0.098} & 0.235 & \best{0.064} & \underline{0.109} & 0.251 & \best{0.177} & \best{0.093} & \best{0.144} & \best{0.182} & \best{0.150}
& $\times$ & 0.108 & 0.202 & \underline{0.145} & 0.893 & 0.337$^{*}$
& \best{0.818} & \best{0.106} & \best{0.946} & \best{0.623}\\
\rowcolor{gray!15}
MASt3R-Fusion (BA only)
& 0.531 & \underline{0.227} & 0.134 & 0.148 & \underline{0.094} & \underline{0.249} & 0.235 & \underline{0.190} & \underline{0.256} & 0.229
& 0.407 & 0.359 & 0.401 & 0.436 & 0.521 & 0.425
& 1.283 & 0.408 & \underline{1.458} & 1.050 \\
DA3-Streaming (DA3-N)$^\ddagger$
& 0.224 & 0.229 & 0.160 & 0.175 & 0.144 & 0.362 & 0.331 & 1.286 & 0.972 & 0.431
& 0.494 & \best{0.066} & \underline{0.120} & 0.170 & \underline{0.117} & \underline{0.193}
& 2.243 & 1.511 & $\times$ & 1.877$^{*}$\\
VGGT(D)-SLAM (DA3-N)$^\ddagger$
& 0.307 & $\times$ & 0.143 & 0.130 & 0.167 & 1.965 & 1.032 & $\times$ & $\times$ & 0.624$^{*}$
& 0.899 & 3.898 & 2.175 & $\times$ & 0.122 & 1.773$^{*}$
& $\times$ & 2.532 & $\times$ & 2.532$^{*}$ \\
\METHODNAME{} (DA3-N)
& \best{0.081} & \best{0.082} & \underline{0.092} & \best{0.052} & \best{0.055} & 0.432 & \underline{0.105} & 0.218 & 0.277 & \underline{0.155}
& \best{0.188} & \underline{0.070} & \best{0.064} & \best{0.127} & \best{0.058} & \best{0.101}
& \underline{1.047} & \underline{0.159} & 1.923 & \underline{1.043} \\
\bottomrule
\end{tabular}
\end{adjustbox}
\par\vspace{2pt}
{\raggedright
\footnotesize
$\times$: diverged run (APE $>$ 5\,m). $^\ddagger$: Sim(3) aligned. $^{*}$Mean over non-diverged sequences only.\par}
\label{tab:slam_eval_all}
\end{table*}

As can be seen in Table \ref{tab:odom_eval_all}, \METHODNAME{} either outperforms or remains competitive with classical methods in translational RPE, and remains comparable in rotational RPE to sub-degree precision.
The table lists RPE results from all three datasets with a mean value reported for each dataset. We note that ORB-SLAM3 and VINS-Fusion diverge on some sequences from the \DATASETNAME{} dataset, which we attribute to poor visual-inertial initialisation. This is not an issue in the ScaRF-SLAM dataset sequences where the device was lifted from a stationary start. 

\METHODNAME{} outperforms MASt3R-Fusion in translational RPE by a significant margin, on all sequences.
Example odometry trajectories are shown in Fig. \ref{fig:traj:odom} which compares MASt3R-Fusion, \METHODNAME{}, and only OKVIS2 out of the classical methods for simplicity.
Runs where different GFMs are used for \METHODNAME{} show a trend of improved performance as model quality increases. Notably, DA3-Nested model (1.4B parameters) shows improved RPE than the runs using DA3-Large (0.3B parameters), a smaller model from the same work. 

\subsection{Pose estimate accuracy as a SLAM system}
To evaluate the global accuracy of \METHODNAME{} as a SLAM system, we enable a basic loop-closure implementation to compare against other methods.
For this evaluation we include both DA3-Streaming, which implements VGGT-Long~\cite{deng_vggt-long_2025} with the DA3 model, and VGGT(D)-SLAM, in which we implement VGGT-SLAM2~\cite{maggio_vggt-slam2_2026} with DA3 models. Both GFM-based methods are excluded from odometry evaluation as they both operate on images only and rely entirely on loop-closures to correct local pose drift.
As their predictions are up-to-scale, we align their estimated trajectories to the reference with a $\mathrm{Sim}(3)$ transformation before computing the APE metric. For all other methods, we fit with a rigid $\mathrm{SE}(3)$ transformation.
We further note that MASt3R-Fusion's implementation applies loop-closure factors as an offline global optimisation equivalent to a bundle adjustment over the entire sequence, which differs from other methods which produce online estimates incrementally. As direct comparisons are therefore not exactly like-for-like, we include their results in a greyed background for reference (denoted BA + LC) and include results from global optimisation without loop-closures (denoted BA only) to demonstrate its influence.
We compute and report the RMSE of absolute pose error (APE) between the estimated SLAM poses and the reference trajectory.

From Table \ref{tab:slam_eval_all} we see that \METHODNAME{} performs well across nearly every sequence, obtaining the best or second-best error for nearly every sequence, and the second-best outlier-aware mean APE across all three datasets. Furthermore, it is competitive with the best classical method tested. 
The table shows APE results where loop-closures are enabled. First, we note that ORB-SLAM3 and VINS-Fusion diverge on the same sequences as in the odometry evaluation. Notably, MASt3R-Fusion outperforms most classical methods on the NewTag (NT) and Christ-Church sequences, which both contain single-loop walks that revisit places rarely and only after long stretches; NT1 in particular is open-loop and does not complete a full closed-loop circuit. These are challenging sequences for all methods, and we attribute part of MASt3R-Fusion's performance to the offline global optimisation: runs without loop-closures that still undergo global bundle adjustment already outperform other methods that apply loop-closures incrementally on several sequences. Visualisations of some runs can be seen in Fig. \ref{fig:traj:slam}, where our SLAM trajectory is compared to DA3-Streaming, OKVIS2, and MASt3R-Fusion.

\begin{table*}[t!]
\centering
\scriptsize
\caption{Reconstruction F-score (\%) $\uparrow$ at 5\,cm on all datasets, with per-dataset means. Note that BigRoom (BR) sequences do not have an aligned ground truth reconstruction.}
\setlength{\tabcolsep}{2.5pt}
\begin{adjustbox}{max width=\textwidth}
\begin{tabular}{l cccccc c ccccc c ccc c}
\toprule
 & \multicolumn{7}{c}{\DATASETNAME{}} & \multicolumn{6}{c}{ScaRF-SLAM} & \multicolumn{4}{c}{Oxford Spires} \\
\cmidrule(lr){2-8} \cmidrule(lr){9-14} \cmidrule(lr){15-18}
 & AT1 & AT2 & NT1 & NT2 & NT3 & NT4 & Mean
 & r01 & r02 & r03 & r04 & r05 & Mean
 & CC02 & CC03 & CC05 & Mean \\
\midrule
\multicolumn{18}{c}{\textbf{Local reconstruction alignment}} \\[1pt]
MASt3R-Fusion
& 56.7 & 46.2 & 49.4 & 55.9 & \underline{54.9} & 51.9 & 52.5
& $\times$ & 63.7 & 60.8 & 68.0 & 42.6 & 58.8$^{*}$
& \underline{41.7} & \underline{38.2} & \underline{40.4} & \underline{40.1} \\
DA3-Streaming (DA3-L)
& 52.6 & 53.4 & 52.4 & 52.6 & 45.4 & 52.3 & 51.5
& 56.7 & 45.0 & 54.1 & 49.6 & 28.8 & 46.8
& 34.8 & 30.4 & $\times$ & 32.6$^{*}$ \\
DA3-Streaming (DA3-N)
& \underline{61.9} & \underline{59.1} & \underline{61.6} & \underline{65.0} & 54.6 & \underline{60.0} & \underline{60.4}
& \underline{74.4} & \best{77.9} & \underline{74.7} & \best{80.5} & \best{73.0} & \best{76.1}
& 32.7 & 34.2 & $\times$ & 33.5$^{*}$ \\
VGGT(D)-SLAM (DA3-L)
& 43.9 & $\times$ & 32.9 & 32.0 & 26.6 & $\dagger$ & 33.9$^{*}$
& 53.3 & 20.6 & 61.9 & 42.2 & 56.8 & 47.0
& $\times$ & 35.8 & $\times$ & 35.8$^{*}$ \\
VGGT(D)-SLAM (DA3-N)
& 50.6 & $\times$ & 31.8 & 42.3 & $\times$ & $\dagger$ & 41.6$^{*}$
& 51.7 & 17.4 & $\times$ & $\times$ & \underline{70.5} & 46.5$^{*}$
& $\times$ & $\times$ & $\times$ & $\times$ \\
\METHODNAME{} (DA3-L)
& 38.1 & 32.9 & 44.6 & 36.6 & 47.1 & 47.4 & 41.1
& 60.5 & 56.4 & 60.3 & 55.7 & 52.8 & 57.2
& $\times$ & 35.2 & $\times$ & 35.2$^{*}$ \\
\METHODNAME{} (VGGT-$\Omega$)
& 45.3 & 51.6 & 46.2 & 53.9 & 49.1 & 50.1 & 49.4
& 53.4 & 57.7 & 60.6 & 59.3 & 63.8 & 58.9
& $\times$ & 31.8 & $\times$ & 31.8$^{*}$ \\
\METHODNAME{} (DA3-N)
& \best{73.0} & \best{66.9} & \best{62.2} & \best{68.4} & \best{64.8} & \best{64.6} & \best{66.7}
& \best{76.2} & \underline{74.4} & \best{75.7} & \underline{72.6} & 66.6 & \underline{73.1}
& \best{54.4} & \best{54.4} & \best{56.0} & \best{54.9} \\
\bottomrule
\end{tabular}
\end{adjustbox}
\par\vspace{2pt}
{
\footnotesize
$\times$: diverged SLAM run (APE $>5$\,m). $\dagger$: out of memory (RAM). $^{*}$Mean over successful sequences only.\par}
\label{tab:recon_eval_all}
\end{table*}
\newsavebox{\recBox}
\newlength{\recW}
\newlength{\recH}
\begin{figure}[t!]
    \centering
    \setlength{\recW}{0.49\columnwidth}
    \sbox{\recBox}{\includegraphics[width=2\recW]{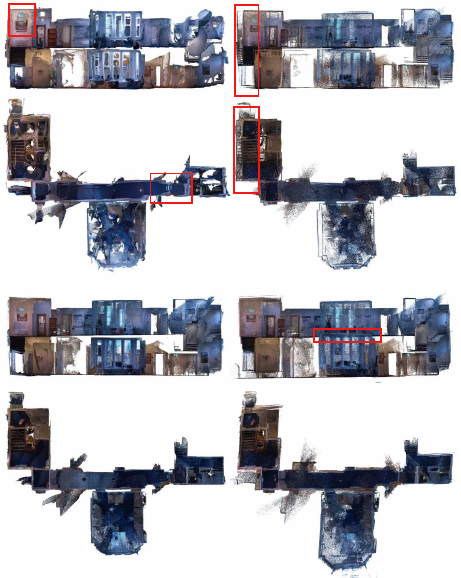}}%
    \setlength{\recH}{0.5\ht\recBox}
    \newcommand{\recQ}[2]{%
        \begin{tikzpicture}
            \clip (#1\recW, #2\recH) rectangle ++(\recW, \recH);
            \node[anchor=south west, inner sep=0] at (0,0) {\usebox{\recBox}};
        \end{tikzpicture}}

    \subfloat[MASt3R-Fusion]{\recQ{0}{1}\label{fig:recon:mast3r}}
    \hfill
    \subfloat[DA3-Streaming (scale 1.105)]{\recQ{1}{1}\label{fig:recon:da3}}

    \vspace{2pt}

    \subfloat[\METHODNAME{}]{\recQ{0}{0}\label{fig:recon:ours}}
    \hfill
    \subfloat[VGGT(D)-SLAM (scale 1.111)]{\recQ{1}{0}\label{fig:recon:vggt}}

    \caption{Side- and top-down views of reconstructions from tested methods on the AT1 sequence from the \DATASETNAME{} dataset. DA3-Streaming and VGGT(D)-SLAM's global fit has a scale of 1.105, 1.111 respectively.}
    \label{fig:recon_comparison}
\end{figure}
\subsection{Reconstruction evaluation}

We follow \cite{zhang_scarf_2026} and evaluate the local accuracy of our dense reconstructions. Given an estimated reconstruction, we divide it into chunks corresponding to 10\,m of trajectory which are fitted to the reference reconstruction with the best-fit transformation and refined with ICP registration. We compute F-score as the harmonic mean of precision and recall, which are respectively computed as the ratio of the number of estimated points to the number of estimated points that lie within a threshold distance to a reference point, and vice versa. In our evaluation we set the valid threshold distance to 5\,cm. For recall, we filter the reference point cloud down to those that lie within 30\,cm of an estimated point in order to filter out unvisited regions.

\begin{figure}[t!]
    \centering

    \begin{tikzpicture}
        \node[anchor=south west, inner sep=0] (img) at (0,0)
        {\includegraphics[width=\columnwidth]{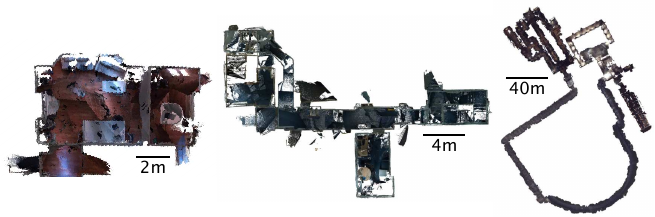}};
    \end{tikzpicture}

    \caption{Global 3D reconstructions from \METHODNAME{} of different sizes from BR2, NT2, and CC05 sequences.}
    \label{fig:recon_example_scale}
\end{figure}

Table \ref{tab:recon_eval_all} shows that in terms of reconstruction quality, \METHODNAME{} attains the best or second-best results for most sequences, closely followed by DA3-Streaming. The table shows the reconstruction evaluation across the three datasets, and similar to local trajectory accuracy, local reconstruction alignment generally improves with model quality across all tested GFM-enabled methods. Particularly, both \METHODNAME{} and DA3-Streaming with DA3-Nested outperform MASt3R-Fusion by a significant margin, and its effects can be seen in Fig. \ref{fig:recon_comparison} which visually demonstrates the difference in local alignment on the global 3D reconstruction. Global reconstructions from other sequences are visualised in Fig. \ref{fig:recon_example_scale}.

\subsection{Real-time performance}
\begin{table}[t!]
\centering
\scriptsize
\caption{Runtime breakdown of \METHODNAME{} (in ms/batch).}
\setlength{\tabcolsep}{3pt}
\begin{adjustbox}{max width=\columnwidth}
\begin{tabular}{l cccccc c}
\toprule
 & Inference & Filter & Rel.\ scale & Loop-closure & Factor graph & Other & Total$^*$\\
\midrule
DA3-L & 725 & 253 & 32 & 90 & 97 & 102 & 1299\\
VGGT-$\Omega$ & 867 & 249 & 36 & 129 & 76 & 436 & 1792\\
DA3-N & 1799 & 239 & 28 & 162 & 63 & 99 & 2390\\
\bottomrule
\end{tabular}
\end{adjustbox}
\par\vspace{2pt}
{\raggedright
\footnotesize
$^*$ Time budget for real-time is 2000\,ms per batch (10 images at 5\,Hz).\par}
\label{tab:runtime}
\end{table}
Table \ref{tab:runtime} shows the processing time of each major component of \METHODNAME{} for three different foundation models. \METHODNAME{} can run real-time using DA3-Large and is near real-time using the DA3-Nested, the largest model tested. Model inference costs around half of the time budget on unoptimised code.
\section{Conclusion}
\label{sec:conclusion}

In this work we introduce \METHODNAME{}, a monocular visual inertial odometry system which fuses general multi-view GFM predictions with IMU data to create odometry estimates and a dense 3D reconstruction that are natively metric in scale. In a factor graph framework, we use a scale consistency factor, which together with the custom up-to-scale GFM pose factor and IMU preintegration observes and tracks true scale. The odometry component of our system achieves tracking accuracy competitive with classical visual-inertial odometry methods and outperforms other GFM-enabled approaches. We later extended the system to support GFM-based loop-closures so as to produce globally accurate reconstructions. 

We evaluated our system on three datasets with different sensor combinations. One key conclusion is that trajectory and reconstruction can be dependent on the particular GFM used, as demonstrated in our use of different models. This allows us to trade off speed and accuracy of different GFM models. Secondly, recent GFMs that learn metric scale from training have improved predictions but are not constrained by physical measurements. They cannot be used as-is for long duration physically accurate reconstruction. For example, \METHODNAME{} outperforms DA3-Streaming (which does not use any direct metric measurements) in accuracy for both trajectory and reconstruction. \METHODNAME{} runs at approximately real-time on a consumer-grade laptop GPU and is mainly time-bound by model inference.

\balance
\small
\bibliographystyle{IEEEtran}
\bibliography{string-short,references}

\end{document}